\documentclass[sigconf]{acmart}

\AtBeginDocument{%
  \providecommand\BibTeX{{%
    \normalfont B\kern-0.5em{\scshape i\kern-0.25em b}\kern-0.8em\TeX}}}

\usepackage{amsmath,amsfonts,bm}

\def\Secref#1{Section~\ref{#1}}

\def\eqref#1{equation~\ref{#1}}

\def\1{\bm{1}}

\DeclareMathAlphabet{\mathsfit}{\encodingdefault}{\sfdefault}{m}{sl}
\SetMathAlphabet{\mathsfit}{bold}{\encodingdefault}{\sfdefault}{bx}{n}

\usepackage{graphicx} 
\usepackage{subcaption} 
\usepackage{xcolor}

\def\Set#1{{\mathcal{#1}}}

\renewcommand{\eqref}[1]{\mbox{Eqn~(\ref{#1})}}

\definecolor{Orange}{RGB}{237,125,49} 
\definecolor{Green}{RGB}{112,173,71} 
\definecolor{Yellow}{RGB}{255,192,0} 

\begin{document}

\title{Bridging the Intent Gap: Knowledge-Enhanced Visual Generation}





\author{Yi Cheng}
\affiliation{%
  \institution{I2R, A*STAR \& National University of Singapore}
  \country{}
  }

\author{Ziwei Xu}
\affiliation{%
  \institution{National University of Singapore}
  \country{}
  }

\author{Dongyun Lin}
\affiliation{%
  \institution{I2R, A*STAR}
  \country{}
  }

\author{Harry Cheng}
\affiliation{%
  \institution{Shandong University}
  \country{}
  }
  
\author{Yongkang Wong}
\affiliation{%
  \institution{National University of Singapore}
  \country{}
  }
  
\author{Ying Sun}
\affiliation{%
  \institution{I2R, A*STAR}
  \country{}
  }
  
\author{Joo Hwee Lim}
\affiliation{%
  \institution{I2R, A*STAR}
  \country{}
  }
  
\author{Mohan Kankanhalli}
\affiliation{%
  \institution{National University of Singapore}
  \country{}
  }
  

\begin{abstract}
For visual content generation, discrepancies between user intentions and the generated content have been a longstanding problem. This discrepancy arises from two main factors. 
First, user intentions are inherently complex, with subtle details not fully captured by input prompts. The absence of such details makes it challenging for generative models to accurately reflect the intended meaning, leading to a mismatch between the desired and generated output.
Second, generative models trained on visual-label pairs lack the comprehensive knowledge to accurately represent all aspects of the input data in their generated outputs. To address these challenges, we propose a knowledge-enhanced iterative refinement framework for visual content generation. We begin by analyzing and identifying the key challenges faced by existing generative models. Then, we introduce various knowledge sources, including human insights, pre-trained models, logic rules, and world knowledge, which can be leveraged to address these challenges. Furthermore, we propose a novel visual generation framework that incorporates a knowledge-based feedback module to iteratively refine the generation process.
This module gradually improves the alignment between the generated content and user intentions. We demonstrate the efficacy of the proposed framework through preliminary results, highlighting the potential of knowledge-enhanced generative models for intention-aligned content generation.
\end{abstract}





\maketitle

\section{Introduction}

\begin{figure}[t]
    \centering
    \includegraphics[width=1.0\columnwidth]{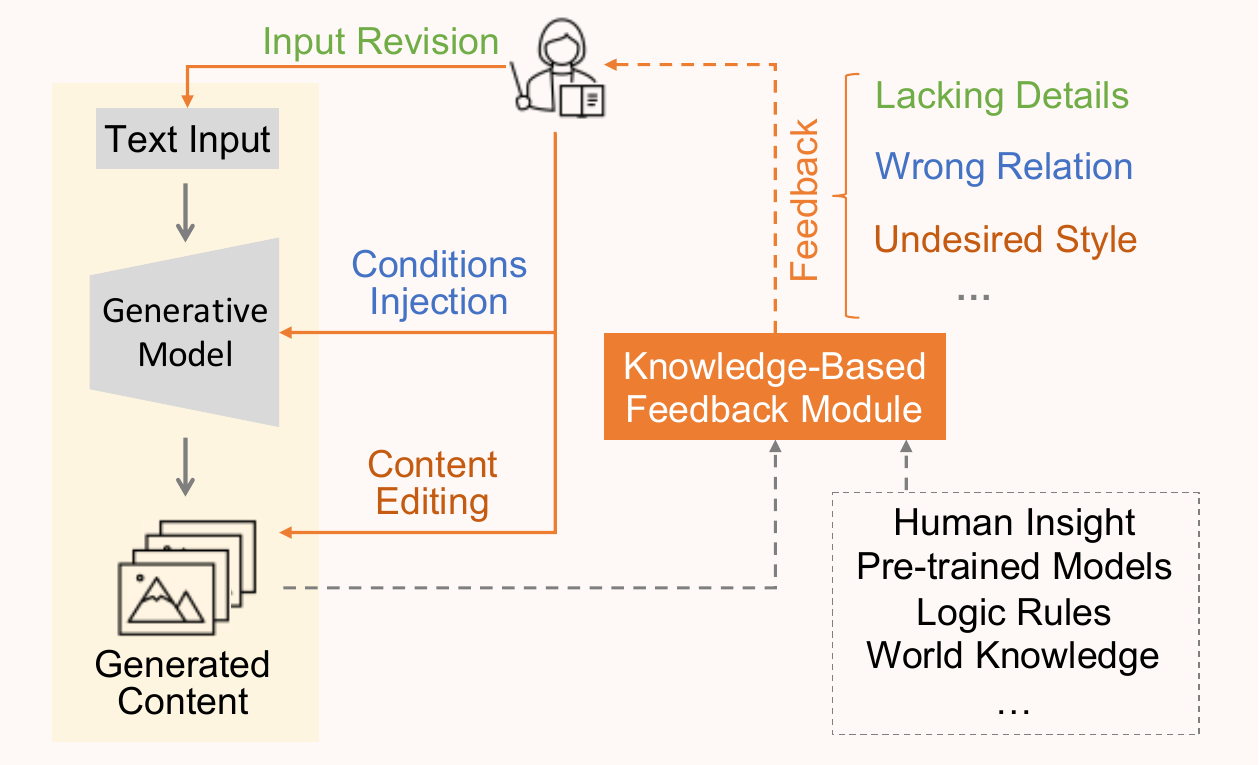}
    \vspace{-1em}
    \caption{
    Our proposed knowledge-enhanced framework for visual content generation. It introduces a feedback loop to the conventional single-round generation process (yellow region). Specifically, a knowledge-based feedback module generates diverse feedback based on the text input and generated content. This feedback is leveraged to iteratively refine the generative process, improving output quality and alignment with user intentions. The conversion of feedback into specific update procedures can be performed either by a human user or automated using pre-defined rules, allowing for varying levels of user involvement.}
    \label{fig:compare_framework}
    \vspace{-1em}
\end{figure}

Visual content generation has gained significant popularity in recent years. Numerous tools and techniques have been developed for generating images and videos, such as DALL·E 2~\cite{Ramesh2022HierarchicalTI}, Stable Diffusion (SD)~\cite{Rombach2021HighResolutionIS} and SORA~\cite{openai2024sora}. 
These advancements enable the generation of high-quality synthetic visual content and have wide applications in various domains, including entertainment, education, and healthcare~\cite{zhang2017compositional,Liu2024SoraAR,singh2021medical}.
Despite these advancements, users often resort to experiment with different prompts, random seeds, or hyper-parameters to achieve the desired result. 
This repetitive process highlights the existence of the \textbf{intent gap}, which remains to be largely hit-and-miss.

\vspace{8pt} 
\noindent\hfill\begin{minipage}{\dimexpr\columnwidth-20pt}
\emph{The intent gap is the discrepancy between the content a user intends to generate and the actual output produced by the generative models, influenced by the limitations of generative models in capturing and interpreting user intentions within a given context.}
\end{minipage}\hfill\null
\vspace{8pt}

\noindent The intent gap arises from two main factors: the inherent complexity of user intentions and the inability of generative models to correctly represent user input. 
User intentions are often implicit, containing subtle details that may not be fully captured by input prompts. This makes it challenging for generative models to accurately reflect the user's intended meaning in the generated content, leading to a mismatch between the desired result and the model's output.
Moreover, generative models~\cite{goodfellow2014generative,Kingma2013AutoEncodingVB,Ho2020DenoisingDP} lack the ability to correctly and fully represent the input in the generated content, primarily due to their data-driven nature. These models are trained on datasets comprised of visual-label pairs, but the labels only provide a fraction of the knowledge required to generate the correct content. The missing knowledge falls into two categories: 1) domain-specific knowledge on the understanding of user preferences, and 2) general knowledge of real-world constraints such as factual information and physical laws. The lack of domain-specific knowledge can result in the generation of undesired content, while the absence of general knowledge can lead to unrealistic or inconsistent outputs. To effectively bridge the intent gap, it is essential to address both factors. This requires a comprehensive strategy that combines iterative refinement to better capture user intentions, with the integration of diverse knowledge sources to supplement the models' existing knowledge base.

Existing works on visual content generation primarily focus on single-round approaches, directly mapping input to the final output. However, this contradicts the iterative nature of visual content generation in real-world scenarios. In such cases, the output is refined through multiple iterations, incorporating feedback until achieving the desired result~\cite{gregor2015draw}. This iterative approach enhances control and feedback integration, effectively bridging the gap between user intention and generated content.
Recent studies have explored iterative refinement techniques to generate the desired visual content, such as using Large Foundation Models (LFMs) to refine text prompts~\cite{Yang2023Idea2ImgIS} or edit generated content~\cite{Wu2023SelfcorrectingLD}.
However, these methods focus on refining either the input or the output, neglecting the challenges arising during the generation process. Moreover, they rely heavily on a single knowledge source, which may not fully capture the diverse knowledge required to effectively bridge the intent gap. 

To bridge the intent gap, it is important to systematically identify and summarize the key challenges faced by generative models. By thoroughly analyzing the limitations, we develop targeted solutions by incorporating diverse knowledge sources to iteratively refine the generated process.
However, integrating various types of knowledge into a comprehensive framework presents considerable challenges~\cite{schuster2022utilizing,zhuang2017challenges}. Knowledge, being inherently diverse, requires careful selection to identify the most relevant information for specific tasks. Furthermore, effectively integrating this knowledge while maintaining accuracy and consistency is crucial for generating high-quality visual content~\cite{hu2018deep}. 

In this paper, we propose a \textbf{knowledge-enhanced iterative refinement framework} for visual content generation, as illustrated in Fig.~\ref{fig:compare_framework}. Our approach aims to address the key challenges faced by generative models across different stages of the generation process.
We begin by summarizing these challenges to provide a comprehensive understanding of the current limitations in visual content generation. To address these challenges, we introduce various knowledge sources at different stages of the generation process. These knowledge sources include human insights, pre-trained models, logic rules and world knowledge.
By iteratively integrating various knowledge sources throughout the generation process, our framework aims to bridge the intent gap and generate user-desired visual content. 
Our preliminary experiments with image generation using diffusion models demonstrate the efficacy of iterative refinement guided by various knowledge.

In summary, the main contributions of this paper are:
\begin{itemize}
\item We define the concept of \textbf{intent gap} in the context of visual content generation and identify the critical challenges faced by generative models that contribute to this gap. We then introduce various knowledge sources to address these challenges.
\item We propose a novel \textbf{knowledge-enhanced iterative refinement framework} to bridge the intent gap in visual content generation. The framework incorporates diverse knowledge sources throughout the generation process to improve the quality and alignment of the generated content with user intentions. The framework can be extended to accommodate new generative models and incorporate additional knowledge sources.
\item We conduct preliminary experiments on image generation using diffusion models to validate the effectiveness of our proposed approach. The results demonstrate that iterative refinement, guided by diverse knowledge sources, significantly improves the quality and alignment of generated visual content with user expectations.
\end{itemize}

The rest of this paper is organized as follows. \Secref{sec:limit} summarizes key challenges faced by generative models and introduces various knowledge sources. \Secref{sec:method} presents our proposed framework. \Secref{sec:exp_res} describes the experimental setup and results. \Secref{sec:conclusion} concludes the paper and discusses future research directions.

\begin{figure}[t]
    \centering
    \vspace{-1em} 
    \includegraphics[width=1.0\columnwidth]{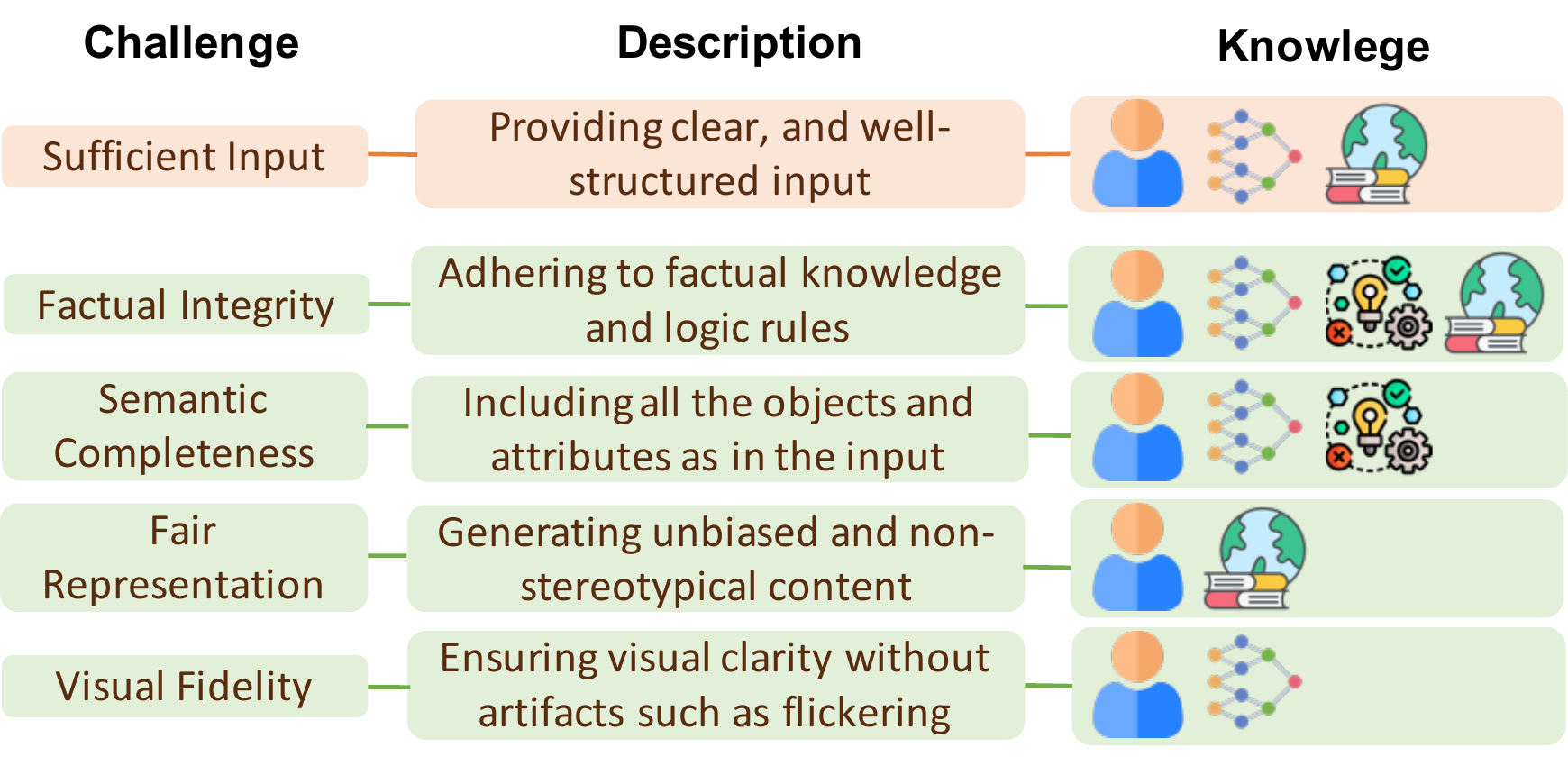}
    \caption{Summary of challenges faced by generative models in visual content generation. 
    Orange denotes challenges related to complexity of user intention, while green denotes challenges related to inability of generative models to accurately represent user input. 
    Each limitation is described along with potential knowledge sources that can help address them. 
    The icons represent different knowledge sources: \includegraphics[height=1em]{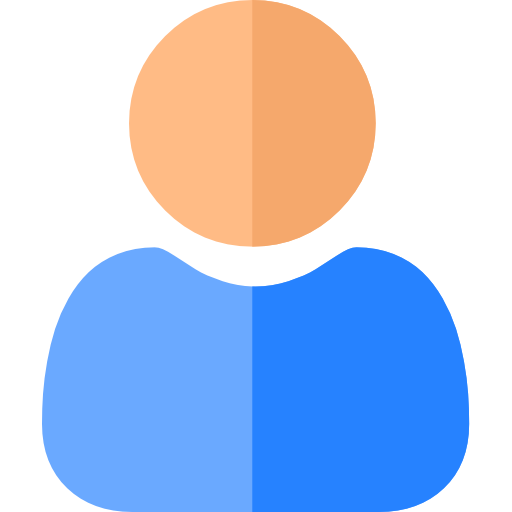} denotes human insight, \includegraphics[height=1em]{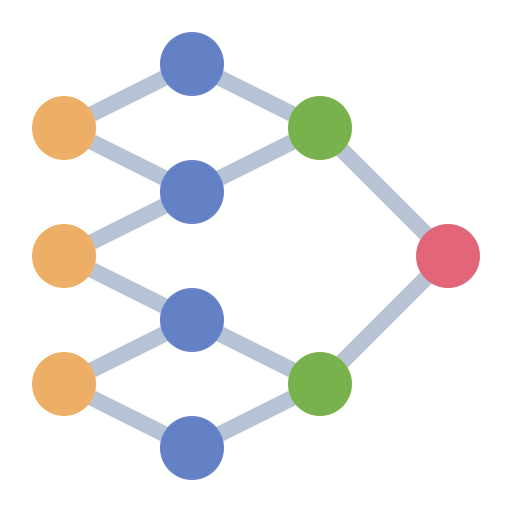} denotes pre-trained models, \includegraphics[height=1em]{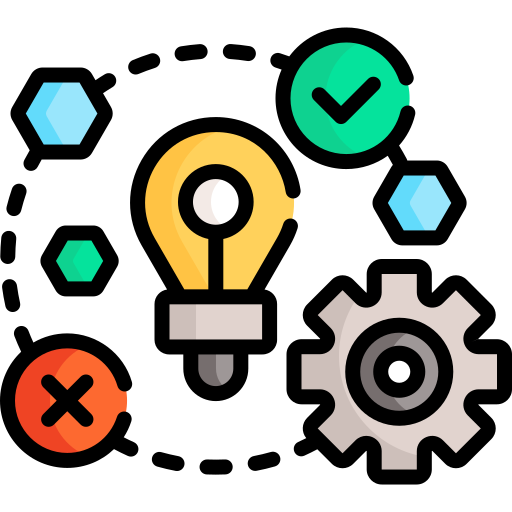} denotes logic rules, and \includegraphics[height=1em]{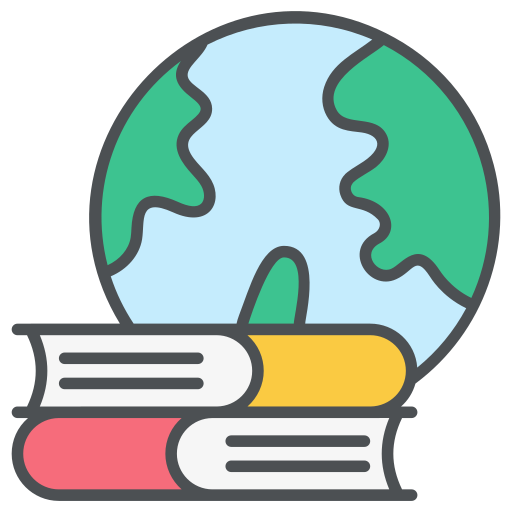} denotes world knowledge.
    }
    \vspace{-2em}
    \label{fig:limitation_summary}
\end{figure}
\section{Challenges and Knowledge}
\label{sec:limit}
This section explores the main challenges faced by generative models in visual content generation, discussing their impact on output quality and highlighting areas for improvement. Additionally, various knowledge sources are introduced to help overcome these challenges and align generated content with user intentions. 
Fig.~\ref{fig:limitation_summary} summarizes the key challenges faced by generative models, where each challenge is described, along with potential knowledge sources for addressing them.

\begin{figure}[t]
    \centering
    \includegraphics[width=0.48\textwidth]{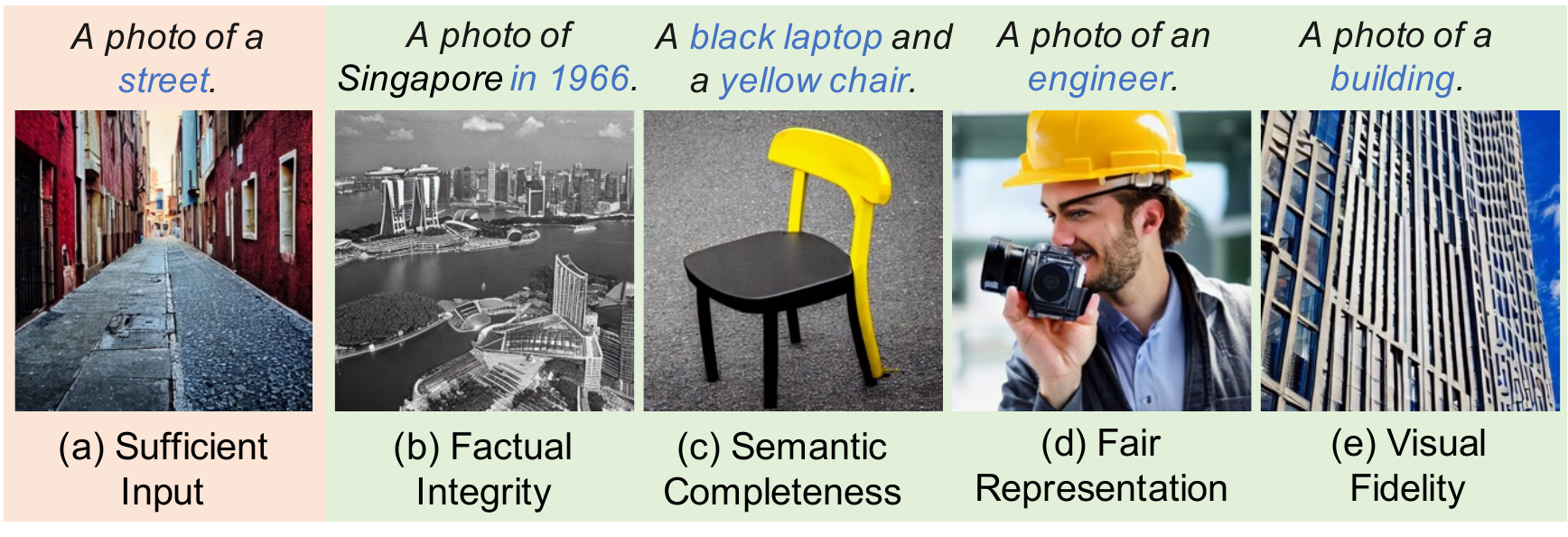}
    \caption{Examples illustrating the challenges faced by generative models. 
    (a) Lacking sufficient input details on the crowd, style, and other characteristics of the street.
    (b) Factual integrity error leading to an anachronistic depiction of Singapore in 1966 featuring modern skyscrapers. (c) Semantic completeness on the flaws in mixing attributes of different objects. (d) Unfair representation of gender: engineers are predominantly depicted as male. (e) Visual fidelity concern with artifacts in buildings, such as skewed lines.}
    \vspace{-1em}
    \label{fig:limitation_examples}
\end{figure}

\subsection{Challenges}
\label{sec:sum_challenges}
The challenges faced by generative models can be broadly categorized into two main areas: 1) challenges related to the input specifications, and 2) challenges related to the inability of generative models to correctly represent user input. Fig.~\ref{fig:limitation_examples} provides examples to illustrate these challenges.

\subsubsection{Sufficient Input}
This refers to when users' intentions are translated into recognizable inputs for the generation model, typically through natural language descriptions. This stage is critical as generative models heavily rely on these inputs to guide the generation process. However, the translation from users' intentions to physical representations is inherently lossy. This would lead to the input data, such as text prompts, failing to fully capture the user's intentions. When the inputs are vague or misaligned with the desired outcome, the model may produce outputs that deviate from users' expectations. For example, the text description ``\textsf{A photo of a street}'' lacks essential details about the street's characteristics, such as crowd density, time of day, and architectural styles, which can lead the generative model to struggle in creating content that accurately reflects the intended scene.

\subsubsection{Factual Integrity}
It refers to the accuracy and consistency of the generated content with respect to factual knowledge and logical rules. It is crucial for ensuring the accuracy, logical coherence, and adherence to the physical laws of the generated content. Maintaining factual integrity poses a significant challenge for generative models. While these models are data-driven, relying on patterns in training data, this approach can lead to a lack of deep understanding of factual knowledge and logical reasoning abilities. Without explicit mechanisms to incorporate factual information and logical rules, generative models may produce outputs that contradict established facts or logical rules.
An illustrative example of this challenge is depicted in Fig.~\ref{fig:limitation_examples} (b), where the image titled ``\textsf{Singapore in 1966}'' inaccurately includes modern skyscrapers, which did not exist at that time, representing a clear factual error.

\subsubsection{Semantic Completeness.} 
It pertains to the model's ability to fully and accurately capture all elements and attributes mentioned in the input, such as the text prompt. This ensures the generated content fully embodies the intended description without missing essential details or inaccurately incorporating attributes. Achieving semantic completeness is challenging for generative models because their training data and model architectures may not provide sufficient information and capacity. Consequently, they may struggle to comprehensively understand and faithfully translate all components and details in the input into the generated output. A clear example of this challenge is illustrated in Fig.~\ref{fig:limitation_examples} (c), where for the input ``\textsf{A black laptop and a yellow chair}'', the model overlooks the laptop while incorrectly assigning its attribute (e.g., ``black'') to the chair, resulting in an inaccurate representation of the input.

\subsubsection{Fair Representation.} 
It refers to the capability of generative models to create content that is fair and unbiased towards different groups and individuals, regardless of their race, gender, age, or other characteristics.
However, achieving fair representation is a significant challenge for generative models.
The key reason for this issue is the lack of diversity and balance in the training data, as it limits the model's exposure to a wide range of social laws and norms. When the training data is skewed towards certain demographics, the model is more likely to learn and reflect those biases in its outputs. This can lead to the reinforcement of stereotypes and the perpetuation of unfair representations. For example, if the training data contains more images of male engineers than female engineers, the model may learn to associate engineering with males, reinforcing gender stereotypes in occupations. Fig.~\ref{fig:limitation_examples} (d) illustrates this problem, where engineers are predominantly portrayed as male, perpetuating unfair gender representation.

\subsubsection{Visual Fidelity.} 
This refers to the ability of generative models to produce visually coherent and aesthetically pleasing content that adheres to general artistic principles and human visual perception. Achieving high visual fidelity requires the generated content to be free from visual artifacts, which can compromise its authenticity and realism.
However, attaining high visual fidelity is challenging for generative models due to the difficulty in accurately modeling and reproducing fine-grained visual details. This arises from the limited ability of generative models to fully grasp and incorporate the knowledge of human visual perception and artistic conventions.
Consequently, generative models may struggle to capture the subtle nuances and aesthetic principles inherent in human-created visual content, resulting in generated images or videos that lack refinement and visual coherence. Fig.~\ref{fig:limitation_examples} (e) exemplifies this problem, showcasing a generated image with skewed edges in buildings.

\subsection{Knowledge Sources}
A variety of knowledge sources, including human insight, pre-trained models, logic rules, and world knowledge, can be leveraged to address the challenges faced by generative models in visual content generation. Each knowledge source addresses specific aspects of the challenges, while collectively functioning in a complete and consistent manner to effectively mitigate issues in existing generative models, as illustrated in Fig.~\ref{fig:limitation_summary}.

\subsubsection{Human Insight}
It refers to the vast information, experiences, and knowledge possessed by users and other individuals within society~\cite{Ouyang2022TrainingLM,Dubois2023AlpacaFarmAS}.
This collective knowledge from users and others represents a valuable resource for addressing the challenges faced by generative models. As users have full knowledge of their intentions, human insight can clarify ambiguities in input and align outputs with personal preferences~\cite{chen2024prompt}. It is also invaluable for assessing the quality of generated content, especially for aspects requiring subjective evaluation. Furthermore, human insight can provide effective update signals based on collected feedback, enabling the model to produce outputs increasingly aligned with human preferences and expectations. 
By leveraging this knowledge source, generative models can better capture users' intentions, adhere to real-world facts, and create outputs that meet expectations regarding semantic completeness, fair representation, and visual fidelity.

\begin{figure*}[t]
    \centering
    \includegraphics[width=0.95\textwidth]{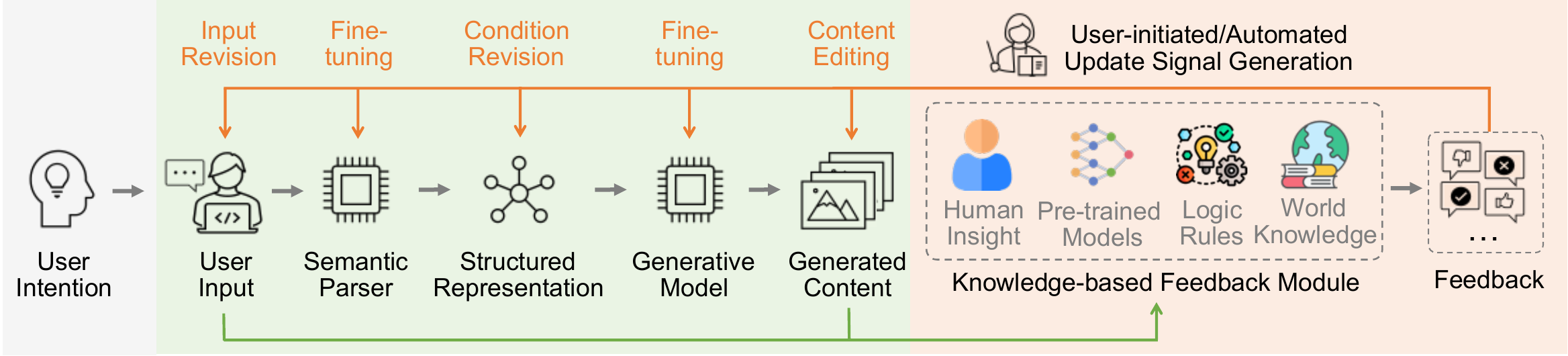}
    \vspace{-1em}
    \caption{
    Overview of the framework. User intention is expressed as user input (gray region). The \textcolor{Green}{Content Generation Baseline} then converts user input into structured representations, which are fed into the generative model to produce content. The \textcolor{Orange}{Knowledge-Based Feedback Module} evaluates the alignment between user input and generated content using knowledge from multiple sources. The feedback is used to enhance the generation process by updating essential components.}
    \vspace{-1em}
    \label{fig:framwork}
\end{figure*}

\subsubsection{Pre-trained Models} 
They act as an important source of implicit knowledge~\cite{Guo2022AUE,Schramowski2021LargePL}. By training on extensive datasets, they can enhance the performance of generative models by providing valuable knowledge and capabilities throughout the generation process. Pre-trained language models like ChatGPT~\cite{Ouyang2022TrainingLM} can help understand and interpret input text prompts, capturing user intent and preferences for generating desired visual content. Pre-trained models can also be used to evaluate the quality and relevance of generated content, providing feedback for refinement. For instance, object detectors~\cite{Minderer2022SimpleOO} can detect spatial incoherence, recognition models~\cite{chen2023ovarnet} can identify wrong attributes, and vision-language models~\cite{Li2023BLIP2BL} can spot factual errors, ensuring greater coherence and consistency. Such feedback serves as an essential signal to refine the generation process. 
Incorporating pre-rained models addresses challenges related to sufficient input, factual integrity, semantic completeness, and visual fidelity in content generation. By enhancing prompt understanding and quality evaluation, pre-trained models ultimately enable the production of high-quality, coherent, and semantically accurate content aligned with user expectations.


\subsubsection{Logic Rules} 
They introduce a layer of explicit and formal rule-based reasoning into AI models~\cite{Xie2019EmbeddingSK, NEURIPS2022_5f96a213,Seo2021ControllingNN,Xu2017ASL}. Unlike the implicit knowledge learned by pre-trained models, logic rules provide a reliable and transparent means of identifying and correcting logical errors in the generated content. 
Generative models may introduce errors during generation, leading to content that violates physical laws or exhibits unrealistic temporal dependencies. These logical errors can be effectively detected by applying logic rules in evaluation. Identifying errors through logic rules provides critical feedback for refining the generation process. This feedback enhances the framework's capability to produce logically coherent content by facilitating updates to essential framework components.
Incorporating logic rules as a knowledge source effectively addresses limitations related to factual integrity and semantic completeness encountered during generation. It offers a systematic approach for identifying and rectifying logical errors, ensuring the generated content adheres to physical laws and temporal dependencies.

\subsubsection{World Knowledge} 
It refers to the comprehensive collection of factual information and real-world context, serving as a highly abstracted form of knowledge about the physical world. 
However, inducing this knowledge through pre-trained models is challenging. Even after large-scale training, models like SORA~\cite{openai2024sora} can produce results that violate physical rules, such as chairs flying in the air. This demonstrates the difficulty in acquiring accurate world knowledge solely from pre-trained models.
Incorporating world knowledge enhances the authenticity of generated content by infusing it with accurate information and contextual details throughout the generation process. Before generation, it enriches input text prompts with relevant contextual information, guiding the process toward historically accurate and culturally appropriate content. During generation, it serves as a reference to ensure the generated content aligns with real-world facts and avoids inconsistencies. The evaluation process can also identify inaccuracies in the generated content by leveraging world knowledge.
By providing a foundation of factual information and real-world context, world knowledge addresses challenges related to sufficient input, factual integrity, and fair representation in content generation, ensuring the generated output is authentic, historically accurate, and contextually appropriate without biases or inconsistencies.

\section{Knowledge-Enhanced Framework}
\label{sec:method}

\subsection{Overall Architecture}
In Section~\ref{sec:sum_challenges}, we explore the challenges faced by generative models during visual content generation. While numerous studies attempts to address these challenges to improve controllability, they predominantly target individual challenges. However, deviation from user intention is a systematic issue arising from a lack of knowledge throughout the generation process.
To address this issue, we introduce a comprehensive framework incorporating multiple knowledge sources throughout the generation process via iterative refinement. As illustrated in Fig.~\ref{fig:framwork}, the framework consists of a content generation baseline and a knowledge-based feedback module.

The generation process begins with the user expressing their intentions as recognizable model inputs, such as natural language description. The Content Generation Baseline then converts user prompts to structured representations, which are fed into the generative model to produce content. The novelty of our framework lies in the Knowledge-based Feedback Module, which takes both the user input and generated content as input. This module evaluates the alignment between the user's intention and the generated content by leveraging various knowledge sources, including human insight, pre-trained models, logic rules, and world knowledge. Based on this evaluation, the module provides feedback on the intent gap. The feedback is then converted into specific update signals, either through user decisions or automated using pre-defined rules. By iteratively refining the process based on these update signals, our framework significantly enhances the alignment of generated content with user intentions. 

We define the content generation process using a model $M$: $\Set{X} \rightarrow \Set{Y}$, where $\Set{X}$ represents the space of user input, and $\Set{Y}$ represents the space of visual content (i.e., images or videos). To incorporate knowledge-based feedback into the generation process, we extend the definition of $M$ as $M: \Set{X} \times \Set{V} \rightarrow \Set{Y}$, where $\Set{V}$ is the space for feedback values. 
We introduce a family of feedback functions $\Set{H} = \{ h : h \in \Set{X} \times \Set{Y} \rightarrow \Set{V} \}$ takes an input $x \in \Set{X}$, the corresponding generated output $\hat{y} \in \Set{Y}$, and returns a feedback value $v_f \in \Set{V}$.
Feedback functions can range from simple binary evaluations to more sophisticated forms, such as natural language feedback.
Feedback value $v_f$ can be consumed by various key components in the framework, such as the user input and the generative model, or can be used to directly edit the output.
As a result, the output of generator $M$ after $k$ rounds of feedback is 
\[
\hat{y}^{(k)} = M(x,v_f^{(k-1)}), \ \text{where} v_f^{(k-1)} = h(x, \hat{y}^{(k-1)}).
\]
We will discuss the feedback mechanism in detail in Section~\ref{sec:feedback}.
The goal of the framework is to align the generated visual content $\hat{y}$ with the input text description $x$ by leveraging feedback functions to assess the quality and relevance of the output. Through iterative refinement based on feedback, the framework aims to improve the alignment between the input and the generated content, ensuring the visual output accurately reflects the user's intentions.

\subsection{Content Generation Baseline}
As illustrated in Fig.~\ref{fig:framwork}, the content generation baseline has three main components: a Semantic Parser, a Structured Representation, and a Generative Model. In the following sections, we detail each component and its possible implementations.

\subsubsection{Semantic Parser.} 
The undesired output from generative models is often linked to their insufficient understanding of complex text descriptions, as seen in systems like CLIP~\cite{Radford2021LearningTV}, which struggle with encoding linguistic structures~\cite{Rassin2022DALLE2IS,Conwell2022TestingRU}. This underscores the critical role of semantic parsers in capturing the nuanced relations within text prompts. By transforming rich and complex language into structured representations, semantic parsers enable generative models to understand user intentions more accurately. The choice of semantic parser can vary, including LLMs such as ChatGPT~\cite{Ouyang2022TrainingLM}, which excel in zero-shot learning and contextual adaptation, converting text descriptions into detailed structured formats. Scene graph parser~\cite{Wu2019UnifiedVE} offers another option to systematically retrieve object attributes, relationships, and overall scenes from text prompts. These structured inputs are vital, providing clear and precise directives to the diffusion model, which forms the basis for generating high-quality visual contents~\cite{Johnson2018ImageGF,Yang2022DiffusionBasedSG,Wu2023ImagineTA,fei2024dysen}. 

\subsubsection{Structured Representations.} Structured representations are vital in translating the complex information of text descriptions into precise inputs for generative models. Text can sometimes be vague or unspecific, as seen in prompts like ``\textsf{A bustling city street at sunset}'', which leaves details like vehicle types or building styles to interpretation. To enhance the specificity and control of the generation process, integrating additional conditions such as object bounding boxes~\cite{Zheng2023LayoutDiffusionCD, Yang2022ReCoRT,Li2023GLIGENOG,Qu2023LayoutLLMT2IEL,Xie2023BoxDiffTS,Xiao2023RBRA}, semantic maps~\cite{Avrahami2022SpaTextSR,Zeng2022SceneComposerAS,Couairon2023ZeroshotSL,Endo2023MaskedAttentionDG}, and poses~\cite{Pham2024CrossviewMD,Cheong2023UPGPTUD,Shen2023AdvancingPI,Fu2023CoPCF,Karras2023DreamPoseFI,Ma2023FollowYP,Qin2023DancingAP} is beneficial. These elements add clarity by defining the spatial placement of objects, delineating areas by their function, and specifying the orientation of figures within the scene. For example, applying a semantic map to the ``\textsf{bustling city street}'' prompt ensures the generated image reflects the desired urban scene accurately. At the same time, pose information can depict pedestrians in dynamic, realistic positions. Incorporating such detailed conditions increases the controllability of the output, ensuring it aligns more closely with user intentions.

\subsubsection{Generative Models.} As the core of our framework, generative models leverage the structured representations derived from text descriptions as input conditions for generating desired visual content.
The selection of generative models is flexible and depends on the defined structural representation of the input.
Depending on the specific requirements and desired output quality, we can opt for training-free generative models~\cite{Chen2023TrainingFreeLC,Mao2023TrainingFreeLT,Phung2023GroundedTS} to maintain the generalizability gained from large-scale pretraining or fine-tune generative models~\cite{Zhang2023AddingCC,Mou2023T2IAdapterLA,Huang2023ComposerCA} based on feedback from initial outputs for enhanced customization and alignment with user intentions. This dual option ensures versatility and effectiveness across various applications and user needs.

\begin{figure*}[t]
    \centering
    \vspace{-1em}
    \includegraphics[width=0.9\textwidth]{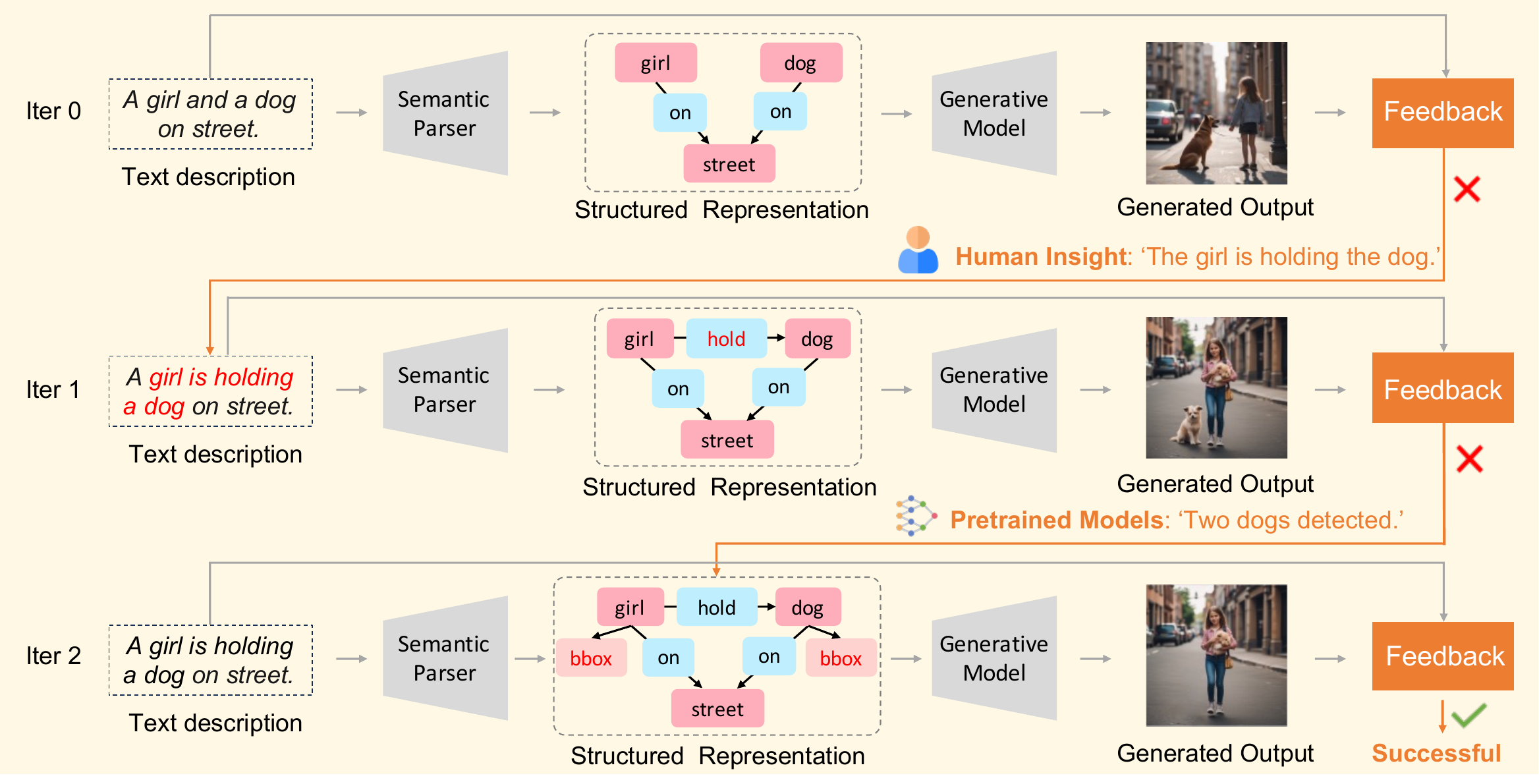}
    \caption{
    An example illustrating the proposed framework for image generation. In the first iteration, the generated content does not meet user intentions due to a vague input description. Human feedback adds details (spatial relation between girl and dog) to the input text, and the image is regenerated. In the second iteration, feedback from foundation models identifies semantic incompleteness (two dogs instead of one). The structured representation is updated with object spatial location using bounding boxes (bbox). After refinement, the model generates an image aligned with the user's intention based on the feedback.}
    \label{fig:framwork_example}
\end{figure*}

\subsection{Knowledge-based Feedback}
~\label{sec:feedback}
To tackle the factors leading to the intent gap, we introduce a knowledge-enhanced iterative refinement framework. This framework is designed to narrow the gap between user intentions and the generated content through three main components: 1) Knowledge Integration in Generation, 2) Update Mechanism, and 3) Iterative Refinement. Fig.~\ref{fig:framwork_example} provides an example of the framework applied to image generation, illustrating its operational structure.

\subsubsection{Knowledge Integration in Generation}
Knowledge-based feedback leverages diverse knowledge sources to provide comprehensive and relevant feedback for refining the generation process. This knowledge can be utilized in two main ways: knowledge-augmented prompting and knowledge-guided generation.

\textbf{Knowledge-augmented prompting} addresses the challenge posed by the inherent complexity of user intentions, which often involve implicit and nuanced details that may not be fully captured by initial input prompts. By incorporating relevant context and preferences derived from various knowledge sources into the prompts, the framework provides more explicit guidance to the generative model.
For example, as shown in Fig.~\ref{fig:framwork_example}, human insight can provide additional knowledge about the relationship between a dog and a girl, addressing the issue of vague input descriptions. 
In addition to human insight, knowledge-augmented prompting can also leverage information from pre-trained models like ChatGPT~\cite{Ouyang2022TrainingLM} or facts retrieved from world knowledge bases. These sources can provide valuable contextual details that enrich the text prompts, enabling the generative model to better understand and represent the user's intentions.
By supplementing the initial input prompts with knowledge from diverse sources, knowledge-augmented prompting effectively addresses the inherent complexity of user intentions. 

\textbf{Knowledge-guided generation} focuses on utilizing diverse knowledge sources to improve the ability of generative models to accurately represent the input in the generated content.  
Generative models often lack essential knowledge, including factual knowledge, logical rules, and social laws, which can lead to generated content containing factual errors, inconsistencies, and biases.

Knowledge can be leveraged to address the challenges in visual content generation by providing two types of feedback. First, it serves as the gold standard for identifying logical or factual errors in the generated content itself. 
Second, knowledge also provides feedback on intent gap by identifying the discrepancy between generated content and text descriptions. Pre-trained models, for instance, can detect inconsistencies or missing information by identifying the relationships and attributes of objects. As illustrated in Fig.~\ref{fig:framwork_example}, pre-trained models can identify the presence of multiple dogs in the generated image, which contradicts the user input of ``a dog''.
Both types of feedback can be in various formats, such as binary indicators or natural language descriptions. 
To convert the feedback into specific update signals, two approaches can be employed. Human users can decide to focus on the important and relevant feedback, or a set of pre-defined rules can be used to minimize the effort required from human users. The details of the update mechanism are further explained in Section~\ref{ssec:update_mechanism}.
To summarize, knowledge-guided generation aims to refine the generative process by identifying errors, ensuring consistency, and embedding essential knowledge. By leveraging the feedback provided by knowledge sources, the quality and relevance of the generated content can be significantly improved.

\subsubsection{Update Mechanism.}
~\label{ssec:update_mechanism}
Upon receiving feedback, our framework employs a comprehensive strategy to refine the content generation through immediate adjustments and model enhancement.

\textbf{Immediate Adjustments} involves directly updating essential components without additional training. It includes refining text descriptions, optimizing structured representations, and adjusting the generated content to align with the desired outcomes. These adjustments are crucial for rapid iterations and significant improvements in the quality of the generated content~\cite{Fernandes2023BridgingTG}.

\textit{Refining text descriptions} aims to clarify ambiguous or vague descriptions in the input. The framework detects areas where the user's intentions may have been misinterpreted. It prompts the user for additional information or suggests alternative phrasing to ensure a more accurate understanding of their requirements.

\textit{Optimizing structured representations} involves adjusting the internal representation of the user's intentions within the framework. These structured representations serve as explicit conditions for the generative models, guiding the content generation process~\cite{tang2024diffuscene}. By refining these representations based on feedback, the framework gains more control over the generation process, producing content that accurately reflects the user's desired outcomes.

\textit{Adjusting generated content} is another critical aspect of immediate adjustments. The framework analyzes the generated content and identifies elements that require modification, such as the overall style, visual attributes of objects, or spatial relationships between elements. By making targeted adjustments through advanced image editing~\cite{Kawar_2023_CVPR,Yang_2023_CVPR} or video editing techniques~\cite{Ceylan_2023_ICCV,Chai_2023_ICCV}, the framework can quickly improve the quality and appeal of the generated content, bringing it closer to the user's expectations.

\begin{table}[t]
\centering
\caption{Performance comparison of our training-free method with state-of-the-art (SOTA) methods, including both training-free and training-based approaches. 
}
\vspace{-1em}
\centerline{
\scalebox{0.9}{
\begin{tabular}{l@{}cccc}\toprule
 & \multicolumn{4}{c}{\textbf{Accuracy}} \\
\cmidrule{2-5}
\textbf{Methods} & Numeracy & Attribute & Spatial & Average \\\midrule
\multicolumn{5}{l}{\textit{Training-free:}}  \\
{\small MultiDiffusion \citep{BarTal2023MultiDiffusionFD}}
& 30\% & 42\% & 36\% & 36\%\\
{\small BackGuid~\cite{Chen2023TrainingFreeLC}}
& 42\% & 36\% & 61\% & 46\% \\
{\small BoxDiff \cite{Xie2023BoxDiffTS}}
& 32\% & 55\% & 62\% & 50\% \\
{\small LMD~\cite{Lian2023LLMgroundedDE}}
& 62\% & 65\% & 79\% & 69\%  \\
{\small \textbf{Ours}}
& 83\% & \textbf{82\%} & \textbf{86\%} & \textbf{84\%} \\
\midrule
\multicolumn{5}{l}{\textit{Training-based:}}\\
{\small GLIGEN~\cite{Li2023GLIGENOG}} 
& 57\% & 57\% & 45\% & 53\% \\
{\small LMD+~\cite{Lian2023LLMgroundedDE}}
& \textbf{84\%} & 79\% & 82\% & 82\% \\
\bottomrule
\end{tabular}
}
}
\vspace{-1em}
\label{tab:exp_quanti}
\end{table}

\textbf{Model Enhancements} focus on updating the trainable components of our framework, specifically the semantic parser and generative models, through retraining with feedback signals. This process incorporates feedback into the models and adapts their underlying algorithms to better understand and process user inputs. 

\textit{Retraining semantic parser} involves using feedback to improve its ability to accurately interpret and understand user intentions. By exposing the parser to a diverse range of user inputs and corresponding feedback, the framework can fine-tune its natural language processing capabilities, enabling it to effectively handle more complex and nuanced user intentions.

\textit{Retraining generative models} involves updating their parameters based on the received feedback. This process allows models to learn from the feedback and adapt their content-generation strategies accordingly. For example, if feedback indicates an unfair representation in the model's output, the retraining process can address this by adjusting the model's algorithms to reduce bias~\cite{shen2023finetuning}. 
By incorporating insights from user preferences and content quality assessments, the generative models can produce more relevant and engaging content that aligns with user expectations.

Our approach offers the flexibility to employ Immediate Adjustments, Model Enhancements, or a combination of both, depending on the specific requirements and nature of the feedback. This strategy forms a dynamic cycle of continuous improvement, empowering our framework to efficiently adapt and respond to feedback.

\subsection{Iterative Refinement}
Iterative refinement aims to progressively align the generated content with user intentions~\cite{bi2024iterative}. This refinement cycle operates through feedback-driven updates until the content meets predefined standards. By bringing human judgment back into the loop, the framework focuses on selecting the most relevant feedback for enhancing the content generation process. This human-in-the-loop approach ensures that the refinements are accurate and appropriate, providing a critical check against the misinterpretation of user intentions or external knowledge.
Each iteration involves applying feedback to enhance the alignment and accuracy of the generated content. The framework ensures focused improvement and efficiency by establishing a limit on iterations or defining specific satisfaction criteria. Through a cycle of generation, feedback collection, and content adjustment, this iterative process progressively narrows the intent gap, ensuring that the final content more precisely reflects the user's original intention.

\begin{table}[t]
\centering
\setlength\tabcolsep{2pt}
\caption{
Ablation results on the contribution of different components. 
``Baseline'' indicates initial results from content generation baseline before refinement, and ``Ours'' denotes the final output after refinement.}
\scalebox{0.9}{
\begin{tabular}{lccc}

\toprule
& \multicolumn{3}{c}{Accuracy} \\
\cmidrule(lr){2-4}
Tasks & Pre-trained SD~\cite{Rombach2021HighResolutionIS} & \textbf{Baseline} & \textbf{Ours} \\\midrule
{\small Numeracy}  & $39\%$ & $\textbf{65\%}$ {\small \textbf{{($\textbf{1.7}\bm\times$)}}} & $\textbf{83\%}$ {\small \textbf{{($\textbf{2.1}\bm\times$)}}}\\
{\small Attribute Binding} & $52\%$ & $\textbf{73\%}$ {\small \textbf{{($\textbf{1.4}\bm\times$)}}} & $\textbf{82\%}$ {\small \textbf{{($\textbf{1.6}\bm\times$)}}}\\
{\small Spatial Relationships} & $28\%$ & $\textbf{72\%}$ {\small \textbf{{($\textbf{2.6}\bm\times$)}}} & $\textbf{86\%}$ {\small \textbf{{($\textbf{3.1}\bm\times$)}}} \\
\midrule
Average & $39.7\%$ & $\textbf{70\%}$ {\small \textbf{{($\textbf{1.8}\bm\times$)}}} & $\textbf{84\%}$ {\small \textbf{{($\textbf{2.1}\bm\times$)}}} \\
\bottomrule
\end{tabular}
}
\vspace{-1em}
\label{tab:exp_abla}
\end{table}
\section{Preliminary Findings}
\label{sec:exp_res}
This section presents the preliminary results of applying our proposed framework to text-to-image generation using diffusion models. However, it is worth noting that the framework can be applied to video generation and use other generative models.

\subsection{Implementation Details}
In our experimental setup, we utilize three knowledge sources to derive feedback: human insight, pre-trained models and logic rules. Human insight is used to design rules for automatically updating structure representations based on feedback, enabling an automatic generation process without human intervention. The pre-trained models include an object detector, Grounding Dino~\cite{Liu2023GroundingDM}, for generating object bounding boxes, and a vision-language model, BLIP-2~\cite{Li2023BLIP2BL}, for recognizing object attributes. Logic rules are employed to define the spatial relationships between objects, such as left, right, above, and below.
For the generative models, we employ a training-free approach for the diffusion models, eliminating the need for additional training. This flexible design allows extensions like retraining diffusion models with feedback and integrating other generative models or additional knowledge sources.


We use the evaluation metric proposed in LMD~\cite{Lian2023LLMgroundedDE} for benchmarking. It is designed for image generation with complex scenarios, including object numeracy, attribute binding, and spatial reasoning.
\textbf{Object numeracy} refers to generating images with a specific number of objects. \textbf{Attribute binding} focuses on correctly assigning attributes to multiple objects. \textbf{Spatial reasoning} evaluates the system's ability to understand and interpret the descriptions of relative object locations.
We generate 100 text prompts per task to test the image-generation capabilities of models. 
For a fair comparison, we follow LMD~\cite{Lian2023LLMgroundedDE} to use a detection-based approach with the OWL-ViT object detector~\cite{Minderer2022SimpleOO} to verify if generated images meet prompt criteria. Task accuracy is determined by the proportion of images that match the prompts.

\subsection{Quantitative Results}
In quantitative analysis, we benchmark our method against several leading text-to-image diffusion models. This comparison spans both training-free approaches (Multidiffusion~\cite{BarTal2023MultiDiffusionFD}, BackGuid~\cite{Chen2023TrainingFreeLC}, BoxDiff~\cite{Xie2023BoxDiffTS}, and LMD~\cite{Lian2023LLMgroundedDE}), as well as training-based methods ( GLIGEN~\cite{Li2023GLIGENOG} and LMD+~\cite{Lian2023LLMgroundedDE}). LMD+ represents an evolution of LMD, incorporating GLIGEN into its framework to enhance spatial control and the use of in-domain instance-annotated data for more precise image generation.

Table~\ref{tab:exp_quanti} illustrates that our method significantly surpasses training-free methods, demonstrating an improvement of over $15\%$. Remarkably, it also exceeds the performance of the training-based method GLIGEN and achieves comparable results to LMD+, a two-stage training-based approach. These results highlight the substantial impact of incorporating feedback mechanisms into the generation process, ensuring that the produced content more accurately reflects user intentions. 
The success of our method highlights the effectiveness of integrating feedback within generative models to improve the relevance and quality of generated images. 

\subsection{Ablation Results}
Table~\ref{tab:exp_abla} showcases our model's remarkable improvements in generation accuracy over the SD~\cite{Rombach2021HighResolutionIS} model across three tasks. Our baseline framework significantly enhances SD's performance, with further gains in our final model. 
For instance, in the Numeracy task, our baseline model improves upon SD's $39\%$ accuracy with a remarkable $65\%$ accuracy, a $1.7$ times increase. Our final model further extends this improvement, achieving an $83\%$ accuracy, marking a $2.1$ times increase over SD. The improvement by baseline showcases the effectiveness of structured representations as input to diffusion models, and the further gain from our final model highlights the effectiveness of our iterative refinement in enhancing content accuracy and aligning with user intentions.

\subsection{Qualitative Results}
\begin{figure}[t]
    \centering
    \includegraphics[width=1.0\columnwidth]{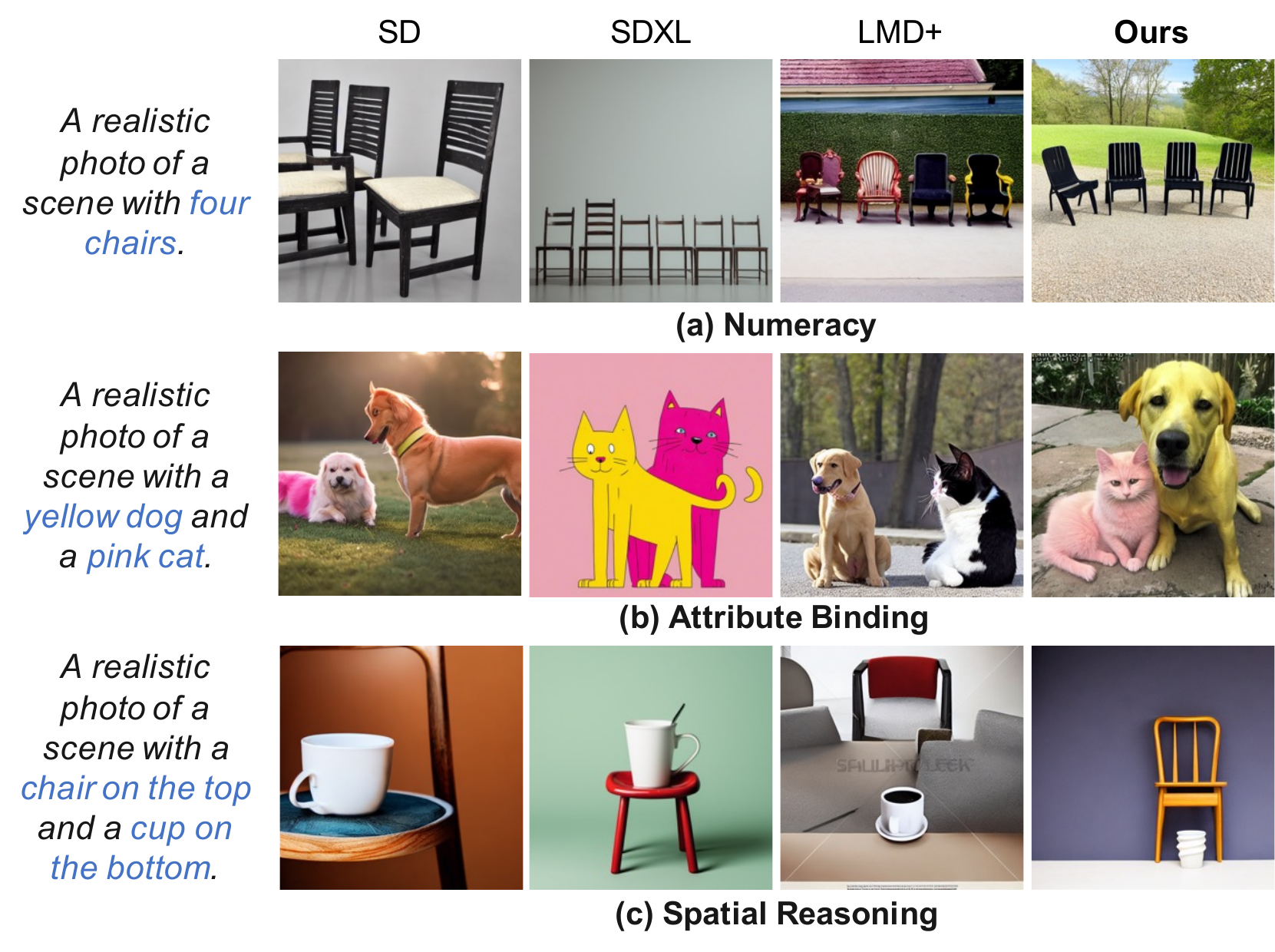}
    \caption{Qualitative comparison between our method and SOTA text-to-image models, including Stable Diffusion~\cite{Rombach2021HighResolutionIS} and LLM-grounded diffusion model LMD+~\cite{Lian2023LLMgroundedDE}.
} 
    \vspace{-1.5em}
    \label{fig:exp_quali}
\end{figure}

In qualitative analysis, we compare our text-to-image generation method with notable existing methods, including SD~\cite{Rombach2021HighResolutionIS}, its enhanced variant SDXL~\cite{Podell2023SDXLIL}, and LLM-grounded Diffusion (LMD)~\cite{Lian2023LLMgroundedDE}, as shown in Fig.\ref{fig:exp_quali}. SD is the baseline due to its robust performance and popularity in text-to-image research. SDXL represents an advancement over SD, offering improved capabilities. LMD introduces a two-stage approach, initially crafting a coarse layout from the textual prompt via LLMs and then integrating individual objects into a unified image according to this layout.

Our analysis reveals distinct advantages of our method in generating accurate and coherent images across various tasks. In the Numeracy task (Fig.~\ref{fig:exp_quali} (a)), while LMD+ manages to render the correct number of chairs, it struggles with consistency in their appearance. This inconsistency stems from LMD+'s reliance on image composition, assembling objects piece by piece into a noise map to form the complete image. Conversely, our approach not only accurately generates the correct quantity of objects but also ensures they share a consistent style and shape.
For the Attribute Binding task (Fig.~\ref{fig:exp_quali} (b)), our method excels in correctly associating attributes, such as color, with the appropriate objects. This contrasts with other methods that sometimes overlook objects or mix up the attributes of different objects.
In the Spatial Reasoning task (Fig.~\ref{fig:exp_quali} (c)), our method demonstrates its capacity to overcome training data biases, successfully generating less common scenarios like a cup placed under a chair. This capability contrasts with other methods that often revert to more predictable or biased configurations, such as placing the cup on top of the chair. This indicates that our method has a deeper understanding of spatial relationships as well as the ability to generate images containing some rare object relations.
These comprehensive results demonstrate our method's superior ability to create images that are not only visually appealing but also more aligned with the intricate details and unique scenarios described in textual prompts.
\section{Conclusion and Future Directions}
\label{sec:conclusion}
This paper introduces a novel framework designed to enhance the alignment of generated visual content with user intentions by addressing multiple challenges faced by generative models. The framework integrates knowledge from various sources, including human insight, pre-trained models, logic rules, and world knowledge, to enable iterative refinement across different phases of content generation. The preliminary results on image generation using diffusion models demonstrate the effectiveness of our framework in bridging the gap between generated content and user intentions. It also highlights the potential of knowledge-enhanced generative models for intention-aligned content generation.

To further advance the field of knowledge-enhanced generative models, future research can explore three key aspects. First, developing methods for automatic knowledge selection and integration is crucial. These methods should aim to minimize human feedback and reduce user fatigue that may arise from systems requiring numerous iterations. Second, exploring techniques to convert knowledge into appropriate representations while ensuring correctness and consistency is essential. This aligns with the emerging direction of knowledge-driven machine learning. Third, extending the knowledge-enhanced generation framework to multi-modal tasks, such as text and audio, is necessary. This extension will enable the production of more coherent and contextually relevant content across different modalities.
Pursuing these directions will lead to sophisticated and reliable knowledge-enhanced generative models, revolutionizing applications by creating accurate, relevant, and user-aligned content.

\begin{acks}
We would like to express our sincere gratitude to Professor Tat-Seng Chua for his insightful suggestions and invaluable comments, which have greatly contributed to the improvement of this work.
\end{acks}

\bibliographystyle{ACM-Reference-Format}
\bibliography{sample-base}

\end{document}